\documentclass[letterpaper, 10 pt, conference]{IEEETran}
\usepackage{graphicx}
\usepackage{courier}
\usepackage{bm}
\usepackage{nccmath}
\usepackage{amssymb}
\usepackage{indentfirst}
\usepackage{algorithm}
\usepackage[pagebackref]{hyperref}
\usepackage{algpseudocode}
\usepackage{orcidlink}
\usepackage{longtable} 
\usepackage{booktabs}
\usepackage[super]{nth}
\usepackage{booktabs,siunitx}
\usepackage[numbers,longnamesfirst,sort&compress]{natbib}
\usepackage{todonotes}
\usepackage{mathtools}
\usepackage{cleveref}

\DeclareMathOperator*{\argmin}{argmin}

\hypersetup{
  colorlinks = true,
  linkcolor = red,
  urlcolor = blue,
  citecolor = green,
  linkbordercolor = {0 0 1}
}

\newcommand{\mat}[1]{\bm{\mathrm{#1}}}

\makeatletter
\newcommand{\mathleft}{\@fleqntrue\@mathmargin0pt}
\newcommand{\mathcenter}{\@fleqnfalse}
\makeatother

\IEEEoverridecommandlockouts
\begin{document}

\title{\LARGE \textbf{DLO-Splatting: Tracking Deformable Linear Objects \\ Using 3D Gaussian Splatting}}

\author{
Holly Dinkel\textsuperscript{*,1}\orcidlink{0000-0002-7510-2066}, 
Marcel Büsching\textsuperscript{*,2}\orcidlink{0000-0001-9296-9166}, 
Alberta Longhini\textsuperscript{2}\orcidlink{0000-0001-9125-6615},
Brian Coltin\textsuperscript{3}\orcidlink{0000-0003-2228-6815},
Trey Smith\textsuperscript{3}\orcidlink{0000-0001-8650-8566} \\
Danica Kragic\textsuperscript{2}\orcidlink{0000-0003-2965-2953},
Mårten Björkman\textsuperscript{2}\orcidlink{0000-0003-0579-3372}, 
Timothy Bretl\textsuperscript{1}\orcidlink{0000-0001-7883-7300}
\vspace{-1em}

\thanks{\textsuperscript{*}Equal Contribution}
\thanks{\textsuperscript{1}Holly Dinkel and Timothy Bretl are with the Department of Aerospace Engineering and the Coordinated Science Laboratory at the University of Illinois Urbana-Champaign, Urbana, IL 61801 USA. e-mail: \texttt{\{hdinkel2, tbretl\}@illinois.edu}}%
\thanks{\textsuperscript{2}Marcel Büsching, Alberta Longhini, Danica Kragic, and Mårten Björkman are with the Division of Robotics, Perception and Learning at KTH Royal Institute of Technology, Stockholm, 114 28 Sweden, e-mail: \texttt{\{busching, albertal, dani, celle\}@kth.se.}}
\thanks{\textsuperscript{3}Brian Coltin and Trey Smith are with the NASA Ames Research Center, Moffett Field, CA, 94035 USA. e-mail: \texttt{\{brian.coltin, trey.smith\}@nasa.gov.}}%
}

\maketitle

\begin{abstract} 

This work presents DLO-Splatting, an algorithm for estimating the 3D shape of Deformable Linear Objects (DLOs) from multi-view RGB images and gripper state information through prediction-update filtering. The DLO-Splatting algorithm uses a position-based dynamics model with shape smoothness and rigidity dampening corrections to predict the object shape. Optimization with a 3D Gaussian Splatting-based rendering loss iteratively renders and refines the prediction to align it with the visual observations in the update step. Initial experiments demonstrate promising results in a knot tying scenario, which is challenging for existing vision-only methods.

\end{abstract}

\section{Introduction}
\label{sec: introduction}

This work presents DLO-Splatting, an algorithm for tracking the shapes of Deformable Linear Objects (DLOs) such as rope for manipulation shape planning and control tasks such as knot tying~\cite{yan2020topological, lagneau2020shapecontrol, yin2021domanipulation, yu2022shapecontrol, jin2022routing}. These tasks are common in applications including robotic surgery, industrial automation, and human habitat maintenance~\cite{lu2016dynamic, viswanath2021disentangling, keipour2022efficient, guo2020algorithm, chang2020manipulation}. Inspired by recent work in cloth shape estimation~\cite{duisterhof2024deformgs, longhini2024clothsplatting}, the DLO-Splatting algorithm approaches the tracking problem using dynamics to predict the state and Gaussian Splatting-inspired rendering  to update the state to track through visually tricky phenomena, such as accurate topology tracking through dense knotting. DLO-Splatting replaces the Graph Neural Network dynamics model used in the prediction step of these previous works with a position-based dynamics model to bypass the reliance on model training and learning. This work makes the following contributions:

\begin{enumerate}
    \item DLO-Splatting synthesizes observations from multiple perspectives to track visually complex topologies which cannot be disambiguated by vision-based tracking alone.
    \item DLO-Splatting uses position-based dynamics with shape smoothness and rigidity dampening corrections to predict the state of the rope one time step into the future without relying on any training data or model learning.
    \item DLO-Splatting uses a 3D Gaussian Splatting-based rendering loss to iteratively refine the predicted state based on new sensor information.
\end{enumerate}

\section{Related Work}

Existing works demonstrate DLO tracking on real data~\cite{xiang2023trackdlo, chi2019occlusion, wang2021cdcpd2, tang2018framework, tang2022track, yang2021dyn, zhang2021llldynamics, schulman2013deformable, ge2014gltp}. These algorithms emphasize vision without physics to track the shape of moving DLOs under various types of occlusion, or otherwise use fiducial markers and physics simulation to ground estimation. Other work extends real-time single-DLO tracking to multi-DLO tracking, estimating the state of multiple DLOs as they are braided while maintaining correct topologies for each object~\cite{xiang2023multidlo}. One of the most challenging aspects of perceiving thin objects, including DLOs, is segmenting them from each other or from the background. 

Existing work on learning unknown object dynamics focuses on revealing properties such as inertial parameters or friction through interaction~\cite{xu2019densephysnet, bianchini2023} while analytical methods using mass-spring or position-based dynamics integrate the shape over time, assuming a small integration time step~\cite{lloyd2007identification, macklin2014pbd, bender2015position,yin2021domanipulation}. Predicting deformable object shapes enables the design of physics-informed state estimators~\cite{schulman2013deformable, tang2017state, longhini2024clothsplatting, abouchakra2024physically,  zhang2024dynamic}.

Vision-based 2D tracking is a well-explored problem with many proposed solutions~\cite{karaev2023cotracker, shi2023videoflow, doersch2023tapvid, trackeverything, teed2020raftrecurrent}. Image-only 2D trackers struggle with occlusions and require additional multi-view or precise depth information to project the results to 3D. Rendering-based losses for tracking have been explored by extensions of NeRF to state estimation~\cite{caporali20233DGSDLO}, dynamic scenes~\cite{mildenhall2021nerf, pumarola2021dnerf, park2021nerfies, li2021NeuralScene, li2023dynibar}, and dynamic extensions of 3D Gaussian Splatting~\cite{kerbl3Dgaussians, dyna3dgs, wu20244dgs, duisterhof2024deformgs}. These methods enable 3D tracking of scene content but often struggle with real-world applicability and long training times, often in the magnitude of hours~\cite{gao2022monocular, buesching2024flowibr}.


\section{The DLO-Splatting Algorithm}

The DLO-Splatting algorithm estimates the 3D state $\hat{\mat{X}}^{t+1} \in \mathbb{R}^{N \times 3}$ of a DLO represented by $N$ nodes over time given an initial state $\mat{X}^0$, the actions applied to the DLO $\mat{a}^{t}$, and visual observations consisting of RGB images $\mathcal{I}^t_k \in \mathbb{R}^{H \times W \times 3}$, where $H$ and $W$ are the camera resolution, from $K$ cameras with camera perspective projection matrices $\mat{P}_k\in \mathbb{R}^{3\times 4}$. The DLO-Splatting algorithm estimates the 3D state of a DLO using a prediction-update framework akin to Bayesian filtering. First, the DLO state is predicted using position-based dynamics (Section \ref{sec: pbd}). This state is iteratively updated using 3D Gaussian Splatting-based rendering (Section \ref{sec: 3dgs}).

\subsection{Prediction with Position-Based Dynamics}
\label{sec: pbd}

Position-Based Dynamics (PBD) derived from physics first-principles are used to predict the next state of the rope given the current estimated state and a gripper action. The position of the grasped node, $\mat{x}_g^t$, is computed as the closest node in the set of nodes describing the shape of the rope, $\mat{X}^t \in \mathbb{R}^{N \times 3}$, to the position of the center of the gripper, $\mat{p}_g^t$, as

\begin{equation}
    \mat{x}_g^t = \argmin_i \| \mat{x}_i^t - \mat{p}_g^t \|.
\end{equation}

\noindent The predicted position of the rope given gripper action $\mat{a}^t = \mat{p}_g^t - \mat{p}_g^{t-1}$ is integrated through Verlet velocity integration of each node in $\mat{X}^t$ for time step $\Delta t$ according to

\begin{equation}
    \mat{X}^{t+1} = \mat{X}^t + \left(\mat{X}^{t} - \mat{X}^{t-1}\right) \Delta t + \frac{1}{2} \mat{F}^t \Delta t^2,
\end{equation}

\noindent where each $\mat{f}_i^t \in \mat{F}^t $ for $\mat{F}^t \in \mathbb{R}^{N \times 3}$ are the summed external forces acting on node $\mat{x}_i^t$ including gravity $\mat{f}_{i,g}^t$, the normal force $\mat{f}_{i,N}^t$, and friction $\mat{f}_{i,f}^t$. Given the constant mass of each node $m_i$ as the total mass of the object $m$ divided by the number of nodes $N$, gravitational constant $g$, and friction coefficient $\mu_f$, the external forces acting on node $i$ are

\begin{equation}
\begin{matrix}
    \mat{f}_{i, g}^t = \left[0, 0, -m_i g \right] \\
    \mat{f}_{i, N}^t = \left[0, 0, m_i g c_i^t\right] \\
    \mat{f}_{i, f}^t = -\mu_f m_i g \mat{v}_{i, xy}^t 
\end{matrix}
\end{equation}

\noindent where binary $c_i^t$ indicates whether node $i$ is in contact with the manipulation plane, i.e., $\mat{x}_{i, z}^t \leq 0$, and $\mat{v}_{i, xy}^t$ is the velocity of node $\mat{x}_i^t$ if $c_i^t = 1$. Furthermore, $\mat{F}^t = \mat{F}_g^t + \mat{F}_N^t + \mat{F}_f^t$.

The rope length is constrained to maintain segment lengths by applying constraint forces, $\mat{n}_i^t \in \mat{N}^t$, as

\begin{equation}
    \mat{X}^t = \mat{X}^t + \mat{\Lambda}^t \mat{N}^t.
\end{equation}

\noindent The parameter $\lambda_i^t \in \mat{\Lambda}^t$ is used to satisfy the constraint on segment length, $L$, given by

\begin{equation}
    \lambda_i^t = \frac{L - \| \mat{x}_{i+1}^t - \mat{x}_i^t \|}{2},
\end{equation}

\noindent and the constraint direction is

\begin{equation}
    n_i^t = \frac{\mat{x}_{i+1}^t - \mat{x}_i^t}{\| \mat{x}_{i+1}^t - \mat{x}_i^t \|}.
\end{equation}

\noindent When the grasped node moves to the new gripper position, the position correction $\mat{x}_g^{t+1} = \mat{p}_g^t + \mat{a}^t \Delta t$ must also be propagated through the rope. After applying Verlet velocity integration, the positions of nodes along the rope are adjusted to keep the segment lengths between adjacent nodes constant. Given lengths $\mat{l}^{t+1}$,

\begin{equation}
    l^{t+1}_i = \| \mat{x}^{t+1}_{i+1} - \mat{x}^{t+1}_{i} \|,
\end{equation}

\noindent and desired length $L$, length corrections, $\Delta \mat{l}^{t+1}$ are computed for each nonzero length in $\mat{l}^{t+1}$ as

\begin{equation}
\Delta l^{t+1}_i = L - l^{t+1}_i \times \frac{\mat{x}^{t+1}_{i+1} - \mat{x}^{t+1}_i}{l^{t+1}_i}
\end{equation}

\noindent and applied to the positions of the nodes on the segment as

\begin{equation}
\begin{matrix}
\mat{x}^{t+1}_i \gets \mat{x}^{t+1}_i - \frac{1}{2}\Delta l^{t+1}_i \\
\mat{x}^{t+1}_{i+1} \gets \mat{x}^{t+1}_{i+1} + \frac{1}{2}\Delta l^{t+1}_i
\end{matrix}.
\end{equation}

\noindent The predicted nodes are finally $\mat{X}_{\text{PBD}}^{t} = \left[ \mat{x}_1^t, \cdots, \mat{x}_N^t\right]^{\intercal}$.

\subsection{Update with 3D Gaussian Splatting-Based Rendering}
\label{sec: 3dgs}

\begin{figure}
    \centering
    \includegraphics[width=0.95\linewidth]{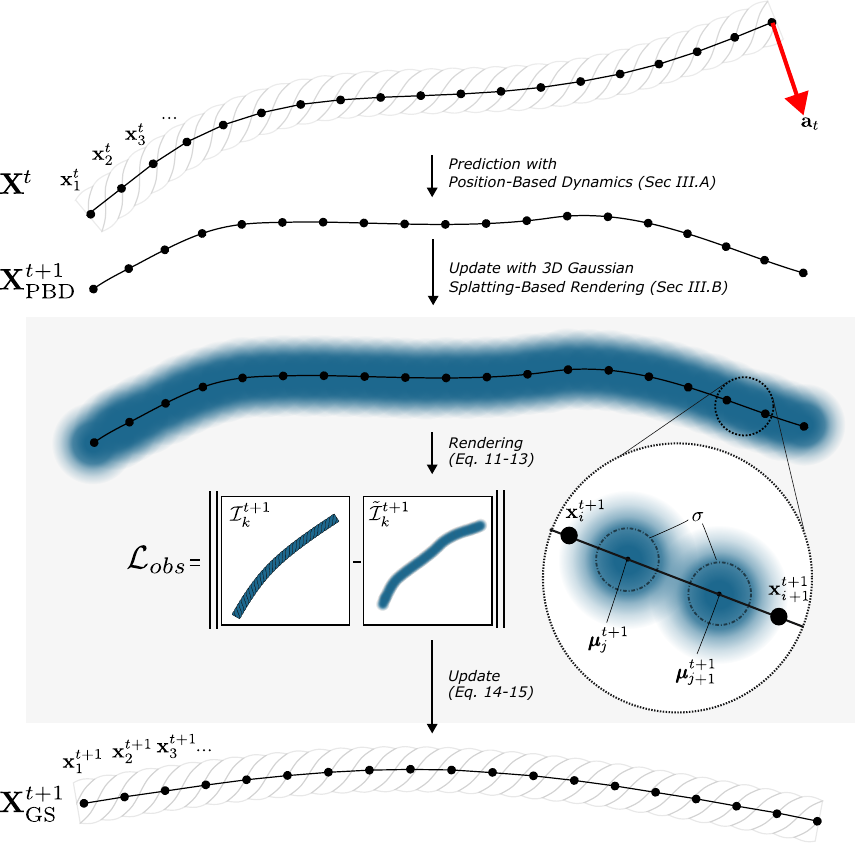}
    \caption{\textbf{The DLO-Splatting Algorithm.} The DLO-Splatting algorithm estimates the 3D state of a DLO using a prediction-update framework akin to Bayesian filtering. The DLO state is predicted using position-based dynamics and is iteratively updated using 3D Gaussian Splatting-based rendering. \vspace{-2em}}
    \label{fig:chain}
\end{figure}

\begin{figure*}
\centering
\vspace{-1em}
\includegraphics[width=\linewidth]{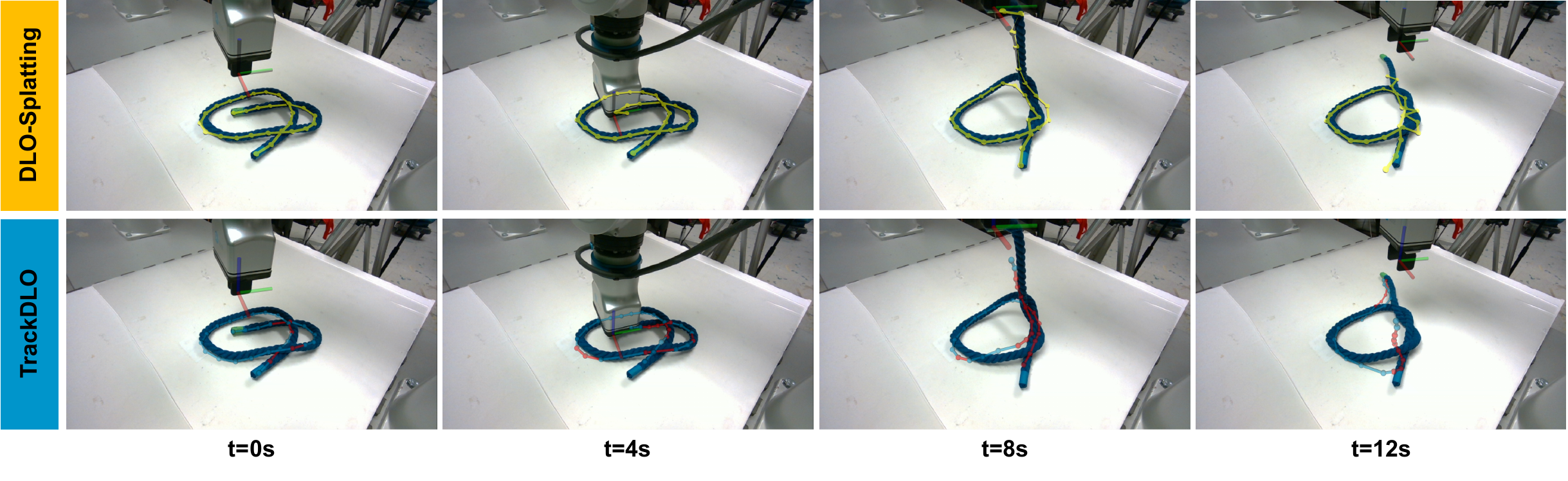}
\caption{\textbf{Qualitative Results.} The DLO-Splatting algorithm is compared to TrackDLO on qualitative tracking during a cross move commonly performed in knot tying. During this move, one tip of the DLO is moved through a loop to create an additional crossing in the topology. After eight seconds, both DLO-Splatting and TrackDLO failed to estimate the correct topology of the DLO, however DLO-Splatting succeeds in tracking the grasped tip. \vspace{-3em}}
\label{fig:demo}
\end{figure*}

In 3D Gaussian Splatting, the scene is represented as a set of $M$ 3D Gaussian distributions, where each distribution $G_j(\mat{\mu}_j, \mat{\Sigma}, \mat{c}_j, o_j)$ is parameterized by its position $\mat{\mu}_j$, covariance $\mat{\Sigma}$, color $\mat{c}_j$,  and opacity $o_j$~\cite{kerbl3Dgaussians}. The 3D Gaussians are distributed on the centerline of the DLO, such that each segment contains $d$ equally spaced Gaussians.
This representation establishes a direct coupling between the 3D Gaussians and the DLO's nodes, allowing the Gaussians positions to deform along with the rope. To capture the shape of the rope, we model the size of all Gaussian with covariance matrix $\mat{\Sigma} = \sigma \mat{1}^{3\times3}$ where $\sigma$ corresponds to the diameter of the rope, resulting in equal-sized spherical Gaussians. Stochastic gradient descent optimizes the Gaussian parameters using the $\mathcal{L}_{obs}$ rendering loss, defined by
\begin{equation}
\label{eq:rendering_loss}
\mathcal{L}_{obs} = \| \mathcal{I}^t_k - \Tilde{\mathcal{I}}^t_k \|_2,
\end{equation}
between ground truth image $\mathcal{I}^t_k$ and rendered image $\Tilde{\mathcal{I}}^t_k$. Rendering $\Tilde{\mathcal{I}}_k^t$ projects the 3D Gaussians onto the image plane of perspective $k$ using $\mat{P}_k$ as

\begin{equation}
    \Tilde{\mathcal{I}}^t_k  = h (\mat{X}_{\text{PBD}}^t, \mat{P}_k) = \left[\mat{c}_j(\mat{u}, G_j) \right]_{\forall \mat{u}\in \mathcal{I}_k^t}.
\end{equation}

\noindent The Gaussians are ordered by their distance to the image plane ($G_0$ is nearest) which allows for their aggregation using $\alpha$-blending. The value of blending factor $\alpha_j$ is estimated for each Gaussian by multiplying its opacity by its Gaussian function, 
\begin{equation}
    \alpha_j(\mat{u}) = o_j  \exp \left(-\frac{1}{2}\left( \mat{u}-\pmb{\mu}'_j \right)^{\intercal}(\mathbf{\Sigma}')^{-1}\left(\mat{u}-\pmb{\mu}'_j \right) \right),
\end{equation}
where $\pmb{\mu}'_j$ and $\mathbf{\Sigma}'_j$ are the 2D projections of the Gaussian parameters onto the image plane. Rendering evaluates for each $\mat{u}\in \mathcal{I}_k^t$ the rendering function given by
\begin{equation}
    \mat{c}(\mat{u}) = \sum_{j \in M} \mat{c}_j \alpha_j(\mat{u}) \prod_{l=0}^{j-1} (1 - \alpha_l(\mat{u})).
\end{equation}
For the state update, DLO-Splatting adapts this static Gaussian Splatting formulation to allow for fast optimization of dynamic rope states. The key difference is DLO-Splatting only models the appearance of the rope with Gaussians, while the geometry of the rope is modeled with the node chain $\mat{X}^t$ as shown in Figure (\ref{fig:chain}). The rendering loss in Eq. \ref{eq:rendering_loss} can be used to iteratively update the node positions as
\begin{equation}
    \hat{\bm{X}}^{t+1} = \bm{X}^{t+1}_{\text{GS}} = 
    \bm{X}^{t+1}_{\text{PBD}} + \Delta \bm{X}^{t+1},
\end{equation}
\begin{equation}
    \text{where} \quad
    \Delta \bm{X}^{t+1} = \argmin_{\Delta \bm{X}} 
    \, \mathcal{L}_{\text{obs}}\left( \bm{X}^{t+1}_{\text{PBD}} + \Delta \bm{X} \right)
\end{equation}
is optimized using Stochastic Gradient Descent (SGD). The original Gaussian Splatting implementation uses direction-dependent spherical harmonics as color representation, enabling the modeling of non-Lambertian effects such as reflections. Since DLOs do not exhibit these visual effects, DLO-Splatting only models their RGB colors instead. The initial RGB color values are the averages of the initial segmentation of the rope. Representing the object thickness with spherical Gaussians also allows for ignoring Gaussian rotations during deformation, simplifying optimization. This work uses \textit{gsplat} as the rasterizer for Gaussian Splatting rendering for improved speed and memory efficiency~\cite{ye2024gsplat}.

\section{Demonstration}

The performance of the DLO-Splatting algorithm is compared to TrackDLO for tracking the state of a rope during a cross move—a common final step in knot tying~\cite{dinkel2024knotdlo, peng2024tiebot}. The demonstration uses three cameras: an Intel RealSense D435, an Intel RealSense D405, and a Luxonis Oak-D Pro, each recording at 30 Hz within a calibrated workspace~\cite{koide2019sthandeye}. An ABB IRB120 robot equipped with an OnRobot 2FG7 gripper manipulates the DLO. Joint states from the robot are recorded at 10 Hz and used to compute forward kinematics in the PyBullet physics simulator to obtain the pose of the grasp center point. The data for the move is saved in a Robot Operating System (ROS) bag file~\cite{ros}. Both methods are initialized using TrackDLO~\cite{xiang2023trackdlo}. As shown in Figure~\ref{fig:demo}, DLO-Splatting more accurately tracks the rope tip as it is lifted out-of-plane by the robot. However, both methods ultimately fail to recover the rope’s final shape and topology.

\section{Conclusion}

This work introduced DLO-Splatting, an in-progress method for estimating the state of a DLO relying on physics and rendering. The DLO-Splatting algorithm is inspired by recent advancements in cloth and DLO state estimation, aiming to handle especially challenging scenarios where vision-only methods often fail. Current limitations include difficulty modeling and constraining self-intersections, a slow update rate of 1 Hz, and reduced performance during occlusions—particularly from the gripper. To address these challenges in the prediction step, future work could use a higher-fidelity simulation to model deformation~\cite{macklin2022warp} or sample physics prediction at a much higher rate to improve resolution of physics constraints. To reduce errors caused by visual artifacts and occlusions, future work may incorporate a Gaussian Splatting-based representation of the entire scene into the rendering process \cite{abouchakra2024physically}. Additionally, the current implementation uses object masking during the update step, and future work could use more recent DLO instance segmentation methods to track multiple objects simultaneously~\cite{zhaole2024robust, dinkel2022RMDLO, zanella2021autogenerated, caporali2022ariadneplus, caporali2022fastdlo, caporali2023rtdlo, viswanath2023handloom}. While DLO-Splatting is still early in development, it demonstrates a promising direction for state estimation in tasks involving complex DLO geometries and dynamics—such as knots—where parts of the object are self-colliding and heavily occluded, making purely vision-based tracking unreliable.

\footnotesize
\section*{Acknowledgments}
\noindent \small The NASA Space Technology Graduate Research Opportunity award 80NSSC21K1292 supported Holly Dinkel and the Wallenberg AI, Autonomous Systems and Software Program (WASP) funded by the Knut and Alice Wallenberg Foundation supported Marcel Büsching. Holly, Marcel, and Alberta thank the 2024 KTH Royal Institute of Technology Robotics, Perception, and Learning Summer School organizing team, including Anna Gautier, Fereidoon Zangeneh, Grace Hung, Jens Lundell, Maciej Wozniak, Miguel Vasco, Olov Andersson, and Patric Jensfelt, for facilitating this collaboration.

\bibliographystyle{iac}
\bibliography{abbreviations,references}

\end{document}